# An Efficient and Small Convolutional Neural Network for Pest Recognition - ExquisiteNet

Shi-Yao Zhou[1*], Chung-Yen Su[1**]

[1]Department of Electrical Engineering, National Taiwan Normal University, Taipei, Taiwan
[*]60875015h@ntnu.edu.tw
[**]scy@ntnu.edu.tw

**Abstract**

Nowadays, due to the rapid population expansion, food shortage has become a critical issue. In order to stabilizing the food source production, preventing crops from being attacked by pests is very important. In generally, farmers use pesticides to kill pests, however, improperly using pesticides will also kill some insects which is beneficial to crops, such as bees. If the number of bees is too few, the supplement of food in the world will be in short. Besides, excessive pesticides will seriously pollute the environment. Accordingly, farmers need a machine which can automatically recognize the pests. Recently, deep learning is popular because its effectiveness in the field of image classification. In this paper, we propose a small and efficient model called ExquisiteNet to complete the task of recognizing the pests and we expect to apply our model on mobile devices. ExquisiteNet mainly consists of two blocks. One is double fusion with squeeze-and-excitation-bottleneck block (DFSEB block), and the other is max feature expansion block (ME block). ExquisiteNet only has 0.98M parameters and its computing speed is very fast almost the same as SqueezeNet. In order to evaluate our model's performance, we test our model on a benchmark pest dataset called IP102. Compared to many state-of-the-art models, such as ResNet101, ShuffleNetV2, MobileNetV3-large and EfficientNet etc., our model achieves higher accuracy, that is, 52.32% on the test set of IP102 without any data augmentation.

## Introduction

Because the number of pests is so many that preventing crops from being attacked by pests is a difficult task for humans. Accordingly, we need a tool having ability to efficiently recognize the pests. Deep learning is a good choice for the purpose. In this section, we will briefly introduce the development of deep learning.

In the last 20 years, deep learning has become very popular in the field of computer vision. However, in fact, deep learning appeared in the world seventy years ago. In 1943, McCulloch and Pitts built the first artificial neural network [1].

However, artificial neural network was difficult to be trained at that time. Accordingly, scientists thought artificial neural network was not useful. The reputation of Artificial neural network was very weak in that period.

Fortunately, in 1986, Rumelhart et al. proposed an effective approach to train artificial neural network, that is, back-propagation [2]. In addition, in order to improve the performance of artificial neural network, the authors proposed to add the sigmoid function into artificial neural network as a non-linear activation.

Since back-propagation and non-linear activation are effective for training the artificial neural network, more and more scientists spent their time on doing related research in the period.

However, scientists found that back-propagation method existed gradient vanishing problem [3]. This hindered the development of deep learning.

In 2011, a new non-linear activation was proposed, that is Rectified Linear Unit activation (ReLU). ReLU activation is said to be able to reduce the effect of gradient vanishing problem. Compared to tanh activation which is widely used before ReLU activation was proposed, ReLU activation is able to decrease much computing time. Nowadays, ReLU activation is widely used in the deep learning model.

In 2012, a famous convolutional neural network called AlexNet took first place in ImageNet Large Scale Visual Recognition Challenge also known as ILSVRC [4]-[5]. It proved that convolutional neural network had great ability for image classification.

From then on, varieties of famous convolutional neural network for image classification sprang up like mushrooms. Such as VGG, ShuffleNet, SqueezeNet, SENet, MoblieNet, GoogleNet, ResNet, DenseNet, efficientNet and so on [6]-[15].

## Related Work

In 2012, Krizhevsky et al. proposed AlexNet [4]. During that period, tanh is a popular activation. However, the authors use Rectified Linear Unit (ReLU) as activation in Alexnet instead. In fact, ReLU activation is better than tanh activation. More importantly, ReLU activation is said to be able to reduce the influence of the gradient vanishing problem.

In 2015, Simonyan and Zisserman proposed a deep learning model called VGG [6]. The authors use two 3x3 kernels to replace one 5x5 kernel. This approach is efficient because the parameters of using two 3x3 kernels is much less than using one 5x5 kernel.

In the same year, Szegedy et al. proposed GoogleNet [11]. GoogleNet is special. The authors concatenated the feature maps produced from different layer and used these concatenated feature maps as the input of next layer. This approach is said to be able to effectively increase the accuracy of image classification. Later, this approach is used in a famous model called DenseNet [13].

In 2016, He et al. proposed ResNet [12]. ResNet was a breakthrough for deep learning model. Although deeper model is beneficial to increase the accuracy of image classification, if the model is too deep, the accuracy will drop, that is, model degradation problem will occur. In order to solve the problem, the authors designed a block called residual block. Nowadays, the residual block has been widely used in varieties of deep learning model.

In 2017, Zhang et al. proposed ShuffleNet [7]. The authors divided the feature maps in the same layer into several groups and made these groups do convolution with different kernels respectively. This approach can decrease the floating-point operations (FLOPs) and the number of parameters in the model.

In the same year, Howard et al. proposed a famous model called MobileNet [10]. The authors split the convolution operation into two steps. First step is the depthwise convolution and second step is the pointwise convolution. This idea is amazing because it can greatly decrease FLOPs and the number of parameters in the model. Nowadays, almost all convolutional layers are replaced by a pointwise convolution and a depthwise convolution.

In the same year, Hu et al. proposed SENet [9]. The authors thought the importance of each feature map in the same layer for increasing the accuracy of image classification is different. Therefore, the authors constructed the SE block.

Huang et al. also proposed DenseNet [13] in 2017. The authors thought that concatenating the feature maps in the previous layer and using them as the input of convolution layer is beneficial to increase the accuracy of image classification.

In 2019, Tan and Le proposed EfficientNet [14]. The authors thought there was three factors influencing the accuracy of image classification. One is image resolution used to train the deep learning model, another is the width of the model, and the other is the depth of the model. Accordingly, the authors provided a formula to teach readers how to adjust the width, depth of model and image resolution to maximize the accuracy of image classification.

In 2020, Han et al. proposed a cheaper operation method to replace the pointwise convolution and named their model as GhostNet [15].

## ExquisiteNet

In this section, we will demonstrate our proposed model architecture and explain the reason why we design this way.

ExquisiteNet mainly consists of two blocks. One is double fusion with squeeze-and-excitation-bottleneck block (DFSEB block), the other is max feature expansion block (ME block) as shown in Fig. 1. The complete architecture of ExquisiteNet is shown in Table I.

*A. DFSEB Block*

In every DFSEB block, we use two residual operations and two SE blocks as shown in Fig. 1(a).

**Double Fusion**. Feature fusion is very effective for increasing the accuracy of image classification. Both DenseNet and ResNet do the feature fusion. However, the feature fusion method of DenseNet and ResNet is very different. DenseNet concatenates the feature maps in all the previous layers to fully fuse the features, on the other hand, ResNet fuses the feature maps in the different layers by adding them up. Unlike ResNet's cheap fusion operation, concatenating the feature maps takes much more computing time. Besides, the operation of concatenating needs large memory size. Accordingly, we choose ResNet's fusion method. In order to fully fuse the features in the shallow layer and the features in the deep layer, we fuse the features two times in every DFSEB Block.

**Non expansion**. According to the paper of ShuffleNetV2 [16], we can learn that if the number of input feature maps and the number of output feature maps of convolutional layer are the same, computing speed will be maximized. Therefore, we don't expand the number of feature maps in our DFSEB Block. According to the paper of MobileNetV2 [17], the inverted residual block is said to be able to effectively increase the accuracy of image classification. However, we found that the inverted residual block is not valid in our experiment.

**SE Bottleneck**. SE block will yeild the weights whose number is the same as the number of previous layer's output feature maps. These weights will multiply the previous layer's output in order to change the importance of every feature map in the previous layer. This approach is said to be able to effectively increase the accuracy of image classification too. In addition, according to the paper of MobileNetV2, using linear bottleneck is better than using non-linear bottleneck. We use this concept in our DFSEB block. To make feature fusion more effective, we use SE block as linear bottleneck in DFSEB block.

*B. ME Block*

We use a maxpooling layer, a pointwise convolutional layer and a batchnormalization layer to construct our ME Block as shown in Fig. 1(b).

Most of the state-of-the-art deep learning models use a depthwise convolutional layer to downsample the feature maps by setting the stride of the layer to 2. In order to increase the computing speed, we use a maxpooling layer instead. Besides, according to our experiment, a maxpooling layer is better than a depthwise convolutional layer with stride set to 2 for pest classification.

In generally, more feature maps in the same layer usually lead to better accuracy. Most of the state-of-the-art deep learning models use a pointwise convolutional layer to expand the number of feature maps. Using a pointwise convolutional layer to expand the number of feature maps is more efficient than using other layers because the operation of pointwise convolution is cheap.

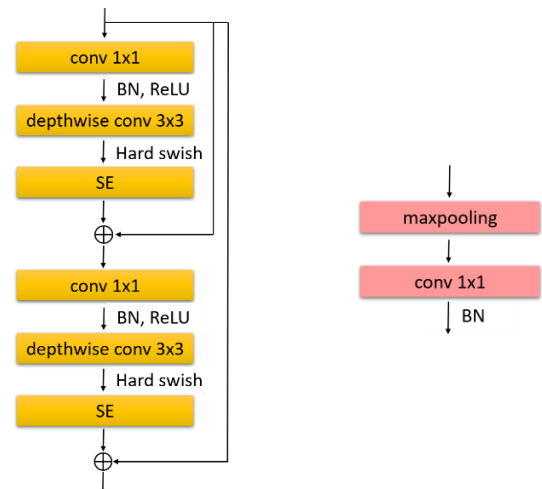

(a) DFSEB block         (b) ME block
Fig. 1. The blocks used in ExquisiteNet

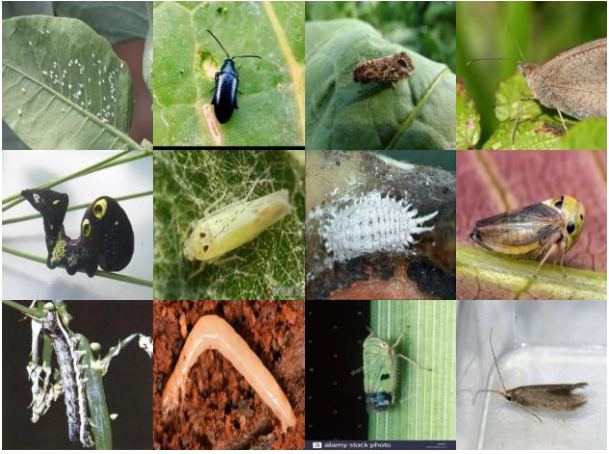

Fig. 2. Several samples in IP102

## Dataset

IP102 is a benchmark pest dataset [19]. There are 102 classes in IP102. IP102 has three subsets: training set, validation set and test set. The number of image in each subset is shown in Table. II. The size of each image is different so images need to be resized to the same shape before we start training the model. Some examples in IP102 are shown in Fig. 2.

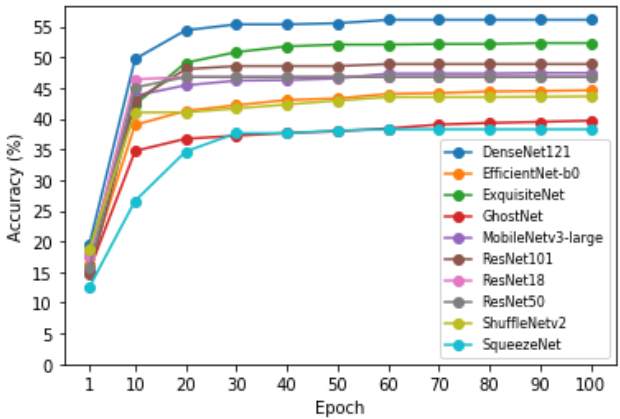

Fig. 3 The accuracy of each model on the test set of IP102

## Experiment and Analysis

### A. Preprocess
All images in IP102 are resized to 224x224 and normalized.

### B. Data Augmentation
We don't augment any image in IP102.

### C. Device and Software Environment for Training
The GPU device we use is GeForce RTX 2080 Ti. Operating System is Ubuntu 16.04. The deep learning tool we used is Pytorch.

### D. Settings
We set batch size to 50 and choose Ranger as optimizer [20]. We set learning rate to 0.001 and don't adjust the learning rate when training the models. Besides, we don't use weight decay method.

### E. Training and Evaluation

We train the state-of-the-art models such as SqueezeNet, ShuffleNetV2, DenseNe121, MoblieNetV3-large, ResNet18, ResNet50, ResNet101, EfficientNet-b0 and our proposed model on the training set of IP102 without pretraining and evaluate the performance of the models on the test set of IP102.

We train every model for 100 epochs and evaluate them at every 10 epochs as shown in Fig. 3. In Fig. 3, the horizontal axis is the epoch being executed, the vertical axis is the best accuracy on the test set of IP102.

### F. Analysis

According to Fig. 3, we can find almost all the models converge at 10th epoch except for SqueezeNet. In addition, we can find DenseNet121 has the best accuracy and our model has next best accuracy on the test set of IP102. However, although SqueezeNet has the fastest computing speed, it has the worst accuracy. In addition, the accuracy of SqueezeNet is much worse than the others.

TABLE I
ARCHITECTURE OF EXQUISITENET

| Input | Operator | Out Channels |
|---|---|---|
| $224^2 \times 3$ | ME | 12 |
| $112^2 \times 12$ | DFSEB | 12 |
| $112^2 \times 12$ | ME | 50 |
| $56^2 \times 50$ | DFSEB | 50 |
| $56^2 \times 50$ | ME | 100 |
| $28^2 \times 100$ | DFSEB | 100 |
| $28^2 \times 100$ | ME | 200 |
| $14^2 \times 200$ | DFSEB | 200 |
| $14^2 \times 200$ | ME | 350 |
| $7^2 \times 350$ | DFSEB | 350 |
| $7^2 \times 350$ | Conv 1×1 | 640 |
| $7^2 \times 640$ | Hard Swish | 640 |
| $7^2 \times 640$ | Averagepooling | 640 |
| $1^2 \times 640$ | Dropout | 640 |
| $1^2 \times 640$ | FC | - |

According to the experimental result of Densenet121, we can learn that reusing the features in the previous layer as the inputs of later layer is an effective approach for classifying the images of pest. Unfortunately, Densenet121 is at the expense of speed, it has the slowest computing speed. Besides, Densenet121 needs a large memory size to be executed. As a result, it is difficult to be used on the mobile devices.

Surprisingly, even though ExquisiteNet only has 0.98M parameters, it gets relatively high accuracy. Alhough the number of parameters in ExquisiteNet is similar to ones in SqueezeNet. The accuracy of ExquisiteNet is 13.6% higher than that of SqueezeNet. Moreover, the computing speed of ExquisiteNet is almost the same as that of SqueezeNet.

The number of parameters, computing speed, accuracy of state-of-the-art models and ExquisiteNet are shown in Table. III.

In addition, only two models reach to the accuracy which is higher than 50% on the test set of IP102. One is DenseNet, and the other is ExquisiteNet.

We are very interested in the region of interest (ROI) when using our model to classify the test set of IP102. In Fig. 4, we show several pictures produced by Grad-CAM method [21]. According to Fig. 4, we can find that when using our proposed model to classify the images, the important region almost perfectly overlaps with the pests. This result is very important because it means that our proposed model maybe have ability to precisely detect the position of the pest in the image.

TABLE II
DETAIL OF IP102

| Subset | Number Of Images |
|---|---|
| Training Set | 45,095 |
| Validation Set | 7,508 |
| Test Set | 22,619 |

TABLE III
MODEL PERFORMANCES ON THE TEST SET OF IP102

| Model | Params (M) | FPS (img/s) | Accuracy (%) |
|---|---|---|---|
| DenseNet121 | 7.05 | 684.80 | 56.11 |
| EfficientNet-b0 | 4.13 | 1031.88 | 44.63 |
| ResNet18 | 11.22 | 1577.33 | 46.85 |
| ResNet50 | 23.71 | 767.52 | 46.79 |
| ResNet101 | 42.70 | 500.53 | 48.90 |
| ShuffleNetV2 | 5.55 | 1686.72 | 43.63 |
| MobileNetV3-Large | 4.33 | 1612.18 | 47.44 |
| GhostNet | 4.03 | 1768.49 | 39.68 |
| SqueezeNet | 0.77 | 1944.88 | 38.26 |
| **ExquisiteNet** | **0.98** | **1936.55** | **52.32** |

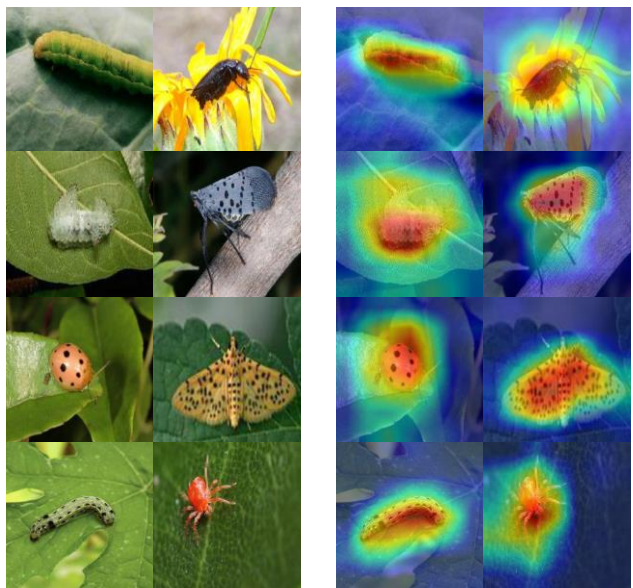

(a) Original      (b) Grad-CAM
Fig. 4 Result Visualization

## Conclusion

Although the performance of the proposed model is good on IP102, we still need to struggle to construct a better deep learning model architecture for the purpose of using ExquisiteNet in real agricultural field. 52.32% accuracy on the test set of IP102 is not good enough.

In order to complete the task of pest detection, we will do our best to design an object detection model by combining our proposed model with one of famous object detection models such as SSD, YOLO, RCNN etc..

Besides, in order to know whether our proposed model can be used in the other fields, we will test our proposed model on several benchmark datasets, such as CIFAR100 and ImageNet in the future.